\documentclass[sigconf]{acmart}

\AtBeginDocument{%
  }

\copyrightyear{2026}
\acmYear{2026}
\setcopyright{cc}
\setcctype{by-nc-nd}
\acmConference[WWW Companion '26]{Companion Proceedings of the ACM Web Conference 2026}{April 13--17, 2026}{Dubai, United Arab Emirates}
\acmBooktitle{Companion Proceedings of the ACM Web Conference 2026 (WWW Companion '26), April 13--17, 2026, Dubai, United Arab Emirates}
\acmPrice{}
\acmDOI{10.1145/3774905.3793124}
\acmISBN{979-8-4007-2308-7/2026/04}

\usepackage{footmisc} 
\usepackage{stfloats} 
\usepackage{subcaption} 

\providecommand{\tightlist}{%
  \setlength{\itemsep}{0pt}\setlength{\parskip}{0pt}}

\begin{document}



\title[Implementation of a Metacognition Framework for Self-Awareness and Self-Regulation \\ in Ensembles of LLMs]{Implementation of a Metacognition Framework for Self-Awareness and Self-Regulation in Ensembles of LLMs}

\author{Charles Courchaine}
\authornote{Both authors contributed equally to this research.}
\affiliation{%
  \institution{Fitchburg State University, National University}
 \city{Portland}
 \state{OR}
 \country{USA}
}
\email{charles@courchaine.dev}

\author{Ricky J. Sethi}
\authornotemark[1]
\affiliation{%
 \institution{Fitchburg State University, Worcester Polytechnic Institute}
 \city{Fitchburg}
 \state{MA}
 \country{USA}
}
\email{rickys@sethi.org}

\author{Hefei Qiu}
\affiliation{%
 \institution{Fitchburg State University}
 \city{Fitchburg}
 \state{MA}
 \country{USA}
}
\email{hqiu@fitchburgstate.edu}

\renewcommand{\shortauthors}{Charles Courchaine*, Ricky J. Sethi*, and Hefei Qiu}

\begin{abstract}
Large Language Models (LLMs) are notorious for struggling with assessing their own uncertainty, detecting knowledge conflicts, or recognizing when problems exceed their expertise; such limitations inevitably undermine reliability and trust in LLMs. In this paper, we present the first implementation\footnote{\url{https://research.sethi.org/metacognition/}\label{fn:DemoLink}} of a \textbf{metacognitive framework} for ensembles of LLMs that addresses these challenges through explicit \textbf{monitoring} and \textbf{control} mechanisms. 

Our system computes a Metacognitive State Vector (MSV) quantifying \textit{self-awareness} for monitoring across five dimensions derived from cognitive psychology: Emotional Response, Correctness Evaluation, Experiential Match, Conflicting Information, and Problem Importance. MSV values also provide \textit{self-regulation} for control, automatically switching between System 1 (fast, single- or multi-node) and System 2 (deliberative, multi-node) processing based on query complexity. For System 2 execution, graph-theoretic algorithms control the assignment of specialized roles (Domain Expert, Critic, Evaluator, Synthesizer, and Generalist) 
to ensemble nodes according to their MSV-quantified metacognitive states.

Our implementation allows users to explore how different query types trigger distinct processing modes. The Proof-of-Concept (PoC) demo showcases the framework with illustrative examples showing appropriate System 1/System 2 routing and helps visualize the metacognitive process via real-time radar charts and decision indicators. 
This PoC implementation demonstrates the feasibility of creating a framework for metacognitive self-awareness and self-regulation in LLM systems.
\end{abstract}





\maketitle

\section{Introduction}

In humans, \textbf{metacognition} can be seen as the dynamic interplay between \textit{\textbf{self-awareness}}, which provides \textbf{monitoring} capabilities, and \textit{\textbf{self-regulation}}, which enables \textbf{control} mechanisms; together, they form an adaptive feedback system 
for
cognition \cite{carver1998self, flavell1979metacognition, nelson1990metamemory, efklides2008metacognition, fleming2014cognitiveneuro}. 

In our previous work \cite{sethi2025llms}, we proposed an analogous metacognition framework for LLMs based on the \textbf{Metacognitive State Vector (MSV)}, which supports both monitoring and control in LLMs; we further incorporated ideas from the Dual-Process Cognitive theory to map System 1/System 2 processing onto ensembles of LLMs mediated by the MSV.
Our main contributions were \textbf{defining} the MSV-based \textit{state-monitoring framework} for allowing \textit{self-awareness} and \textbf{designing} a complementary \textit{control architecture} that enables  
\textit{self-regulation}.

This kind of metacognitive framework is crucial for addressing LLMs' inability to assess their own uncertainty \cite{steyvers2025metacognition} and their tendency to systematically generate hallucinated content \cite{Huang2025hallucination}. By operationalizing explicit monitoring and control mechanisms through the MSV, our approach directly targetted these metacognitive deficits that undermine current LLMs' reliability and trustworthiness.

\subsection{Contributions}
In this work, we now \textbf{implement} a proof-of-concept (PoC) of the entire system above, as shown in Figure \ref{fig:control-flow-process}. We provide the full codebase, installation video, and demo video at the above link\footref{fn:DemoLink} and present the following contributions:

\begin{itemize}
\item \textbf{Implementation of the MSV-based Metacognition Framework}, including  facilitating automatic System 1/2 switching and role assignment to nodes in ensembles of LLMs
\item \textbf{Functional user interface} demo, including: Query Input, MSV Radar Charts, System 1/System 2 decision indicator with threshold-based explanation, network graph visualization of node role assignments, and expandable interface for node contributions and synthesis.
\item \textbf{Illustrative examples} showing the framework's capabilities through qualitative demonstrations of different system behaviors across query types with varying metacognitive complexity
\end{itemize}

\begin{table*}[t]
\centering
\begin{tabular}{|p{3.4cm}|p{10cm}|p{3.3cm}|}
\hline
\textbf{Dimension} & \textbf{Formula} & \textbf{Meaning} \\
\hline
\textbf{Emotional Response} & 
$ER  = \bigl(ER_v,\, ER_a,\, \vec{\sigma}\bigr)$, with $ER_v = \sum_{i=1}^{n} \epsilon_i v_i$, with $\epsilon_i \geq 0$, $\sum_{i=1}^{n} \epsilon_i = 1$ 
& Valence, affect, \& stability \\
\hline
\textbf{Correctness Evaluation} &
$CE = \alpha_1 F_1(\text{logical\_consist}) + \alpha_2 F_2(\text{factual\_acc}) + \alpha_3 F_3(\text{contextual\_approp})$ 
& Logical, factual validity \\
\hline
\textbf{Experiential Matching} &
$EM = \omega_1 K(\text{resp, know\_base}) + \omega_2 H(\text{resp, hist\_resp}) + \omega_3 C(\text{prompt, cue\_famil})$ 
& Familiarity with past cases \\
\hline
\textbf{Conflicting Information} &
$CI = \delta_1 D(\text{internal\_consist}) + \delta_2 D(\text{source\_agree}) + \delta_3 D(\text{temporal\_stabil})$ 
& Contradiction detection \\
\hline
\textbf{Problem Importance} &
$PI = \beta_1 C(\text{potential\_conseq}) + \beta_2 U(\text{temporal\_urg}) + \beta_3 I(\text{scope\_impact})$ 
& Task priority \& scope \\
\hline
\end{tabular}
\caption{The Metacognitive State Vector (MSV)}
\label{table:msv}
\end{table*}

\section{Metacognition Framework Implementation}

Central to our metacognition framework is the \textbf{Metacognitive State Vector (MSV)}, a 5-dimensional vector that quantifies metacognition on a common scale, with explicit formulas for each component, as seen below and in Table \ref{table:msv}:

\begin{itemize}
    \item \textbf{ER (Emotional Response)}: Aggregates affective states as intensities of multiple emotion categories, each weighted by its contextual importance in order to represent the computational analogue of affective metacognitive experiences.
    \item \textbf{CE (Correctness Evaluation)}: Weighted composite of logical consistency, factual accuracy, and contextual appropriateness; forms an \textit{uncertainty signal} $= 1 - CE$ that triggers deeper analysis.
    \item \textbf{EM (Experiential Matching)}: Measures similarity of the current response to prior knowledge and experience; yields \textit{unfamiliarity signal} $= 1 - EM$ to promote exploration when novelty is detected.
    \item \textbf{CI (Conflicting Information)}: Degree of internal, source-level, or temporal inconsistency that flags potential contradictions requiring higher-order deliberation.
    \item \textbf{PI (Problem Importance)}: Evaluates the potential consequences, urgency, and scope of a problem to prioritize cognitive and computational resources toward higher-impact or time-sensitive tasks.
\end{itemize}

In the PoC, these are self-reported and single-channel validation of state-of-the-art approaches for each dimension are relegated to future work.

\subsection{Dual-Process Mapping Implementation}
\begin{figure}[b]
    \centering
    \includegraphics[width=\linewidth]{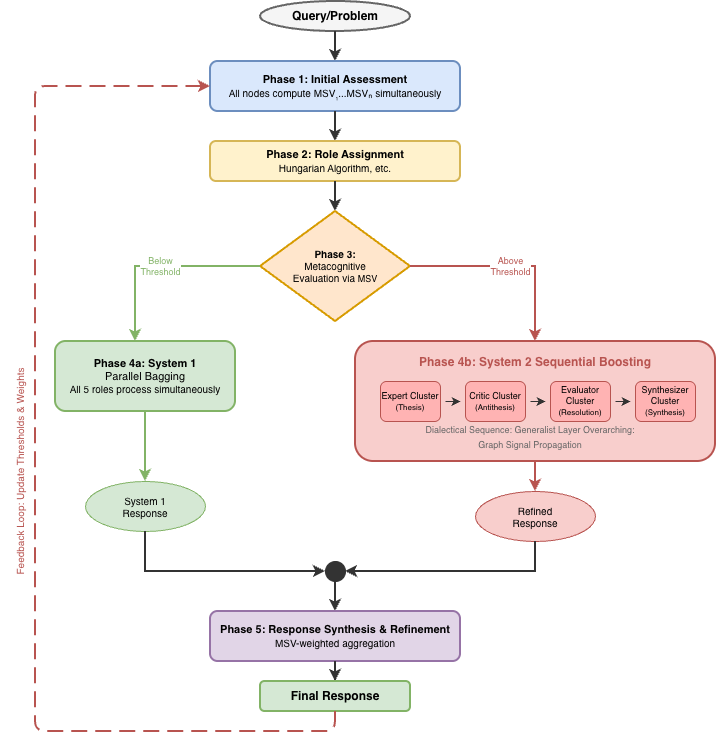}
    \caption{5-phase Ensemble Control Flow Process.}
    \label{fig:control-flow-process}
\end{figure}

We map ensembles of LLMs to either System 1 (fast, low-cost, bagged ensemble or single node) or System 2 (slow, deliberative, boosted ensemble). The MSV values govern \textit{when to escalate} from System 1 to System 2.
The control flow for deciding between a System 1 output or a deeper System 2 output operates through a five-phase orchestration protocol as seen in Figure 
\ref{fig:control-flow-process}, where we also see a feedback loop in anticipation of extension to meta-reasoning and which also underlies computational thinking \cite{sethi2020essential, russell1991principles}:

\begin{enumerate}
\def\labelenumi{\arabic{enumi}.}
\tightlist
\item
  Parallel MSV computation where each node independently evaluates the
  query to generate its own metacognitive assessment MSV;
\item
  Role assignment using a conflict resolution mechanism (e.g., Hungarian
  algorithm) to ensure role diversity while considering
  MSV-derived fitness scores;

  
\item
  System mode selection: aggregated MSV values are compared against
  configurable thresholds to determine System 1 (fast, parallel) versus
  System 2 (slow, sequential) activation;
\item
  Query execution following either parallel bagging (System 1) or
  sequential boosting (System 2) patterns; and
\item
  Final response synthesis with MSV-weighted aggregation; also handles early-stopping verification and formatting.
\end{enumerate}

The MSV aggregation can employ complex strategies (like boosting) but initially uses:

\begin{enumerate}
\def\labelenumi{\arabic{enumi}.}
\tightlist
\item
  arithmetic mean for System 1 decisions
  ($\text{MSV}_\text{avg} = \frac{1}{n}\sum{\text{MSV}_i}$), 
\item
  weighted average by confidence for System 2
  $\text{MSV}_\text{weighted} = \sum{(\text{CE}_i \cdot \text{MSV}_i)}/\sum{\text{CE}_i}$), and 
\item
  percentile-based aggregation for conflict detection 
  (e.g., using the 75th percentile of CI values as a control signal to trigger  more deliberation or synthesis prompts that surfaces disagreement rather than forcing false consensus).
\end{enumerate}

The System 2 sequential execution can specifically order roles, e.g., as
Domain Expert $\rightarrow$ Critic $\rightarrow$ Evaluator $\rightarrow$ Synthesizer, with each node's
output enriching the context for subsequent processing and early
stopping triggers, e.g., when consecutive MSV confidence scores might
exceed 85\% or conflict scores drop, say, below 20\%.

\subsection{Graph-Theoretic Control Layer}

In our framework, nodes (LLMs/agents) assume roles (Domain Expert, Critic, Evaluator, etc.); \textit{role transitions} are driven by MSV-weighted softmax policies and edge activations (sigmoid with temperature).

The graph-theoretic control system is implemented as a directed graph
G(V, E, W) where the vertices V represent individual LLM nodes
, edges E denote communication
pathways between nodes, and edge weights $W(e) = (w(e), \mu(e)(M))$ 
combine \textit{static} base weights ($w(e)$)  with \textit{dynamic} metacognitive transition
functions ($\mu(e)(M)$).

The base weights are the fixed, pre-determined weight that represents the inherent connection strength between two nodes, regardless of their current state, and can be initialized based on things like the topology structure (e.g., 1.0 for direct neighbors or 0.5 for distant neighbors) or node similarity (e.g., two Expert nodes might have w=0.8) or historical preference (this would be closest to biological neurons where neurons that fire together, wire together; e.g., Node1 $\rightarrow$ Node2 historically produces good results, etc.) or perhaps manually based on use-case context or expert domain knowledge (e.g., we always want Critics to influence Evaluators strongly or some such). 

The dynamic weights change in real-time based on the source node's current metacognitive state and can be calculated each time based on current MSV as a weighted sum (as seen below). The final edge weight currently combines both as multiplicative (but could be additive or weighted, as well). 

Each node maintains a local Metacognitive State Vector (MSV) computed
across the five dimensions (ER, CE, EM, CI, PI) and can dynamically
assume roles from the set R = \{Domain Expert, Critic, Evaluator,
Synthesizer, Generalist\}.

The edge activation function is

$\mu(e_{ij})(M_i) = \sigma(\alpha_1 \cdot ER_i + \alpha_2 \cdot CE_i + \alpha_3 \cdot EM_i + \alpha_4 \cdot PI_i + \alpha_5 \cdot CI_i)$
or, more generally:
$\mu(e_{ij})(M_i) = \sigma(\sum{\alpha_k \cdot \text{MSV}_k})$ 
with:
$\sigma(x) = \frac{1}{1 + e^{-x/\tau}}$
where the sigmoid uses a temperature parameter $\tau$ to modulate information flow based on the source node's metacognitive state. Role transition probabilities are calculated using $P(r' | r, M) = softmax(T(r, r', M))$, where

$T(r, r', M_i) = w_1 \cdot ER_i + w_2 \cdot CE_i + w_3 \cdot EM_i + w_4 \cdot PI_i + w_5 \cdot CI_i $
or, more generally:
$T(r, r', M_i) = \sum{w_k \cdot \text{MSV}_k} $

represents the transition score from current role $r$ to target role $r'$,
with role-specific weight vectors $w$ learned or manually configured
(e.g., Critic→Expert transition weights CE heavily at 0.4 while CI at
0.1, reflecting that high confidence triggers expert consultation). The
system persists state history to enable continuous learning (can update base weights or dynamic weights with time) and maintains both synchronous (for System 2 sequential boosting deliberation) and asynchronous (for System 1 processing via parallel bagging) execution paths.

The main thing is that the source node’s metacognitive state
determines how strongly it “broadcasts” to its neighbors; e.g., a
node with high uncertainty might reduce its outgoing edge weights
(quieter voice) but a node with high confidence and low conflict
might increase them (louder voice), etc.

\textit{Resolving Role Conflicts}: Suppose both Node 1 and Node 3 want to be Expert \textbf{and}
Node 2 and Node 4 both prefer the Critic role. The Hungarian algorithm
solves this as an assignment problem and, in larger ensembles (or
perhaps more importantly in smaller ones), the conflict resolution
ensures we don't end up with situations like five Critics and no Expert, or five Experts
with no one to challenge their assumptions, etc. In addition, the ensemble itself can be assigned a role, or a  distribution of final role assignments within the ensemble, showing how it ``thinks'' overall.

\textit{Scalability and Role-Based Clustering}. \label{beyond-5-nodes}
For ensembles beyond 5 nodes, the framework supports hierarchical organization where nodes with the same role assignment can naturally form functional clusters. At scale, the Hungarian algorithm's role assignments can produce multiple Critics, multiple Experts, etc., which then operate as coordinated subgroups. E.g., high CI scores (>70) might result in more Critic role assignments, effectively forming a "critic caucus" that engages in intra-cluster deliberation before presenting unified challenges to Expert clusters. High EM scores (>80) across multiple nodes might similarly create collaborative synthesis clusters where confident nodes reinforce each other's contributions through the existing edge activation mechanism. High PI scores (>75) could trigger increased Evaluator role assignments, establishing hierarchical validation structures where Expert clusters propose and Evaluator clusters verify. Importantly, this role-based clustering operates within the stable graph topology established during initialization; i.e., the structure remains fixed while activation patterns and role distributions adapt to metacognitive demands. This separation of stable structure from dynamic role assignment preserves learning continuity while enabling emergent functional organization at scale.
The inter-role flow remains a fixed sequential pipeline while the graph-theoretic machinery operates within intra-role clusters as an overlay on stable base topology, as seen in Figure \ref{fig:control-flow-process}.

\subsection{Implementation Details}
The framework is implemented in Python and consists of a codebase of \~{}1,500 LOC, publicly available at the above link\footref{fn:DemoLink}. All LLM calls are made to llama 3.2 via ollama and all MSV visualizations are done utilizing FastAPI, HTMX, and Bokeh. 
Hardware requirements are also simple: a Macbook M3 Pro with 18GB of RAM for this initial implementation with possible secondary farming out of calls to the Google Cloud infrastructure.

Our system can optionally employ a hybrid architecture where System 1 queries execute 
locally via ollama for minimal latency, while System 2 deliberations can 
then distribute node execution across Google Cloud Vertex AI instances, 
enabling parallel processing of multiple roles (e.g., Critic, Evaluator, 
Synthesizer running concurrently). This hybrid approach would balance response 
time with resource constraints: simple queries complete locally but 
complex multi-node deliberations can then leverage cloud parallelization to 
reduce System 2 processing.

\begin{figure}[t]
    \centering
    \includegraphics[width=\linewidth]{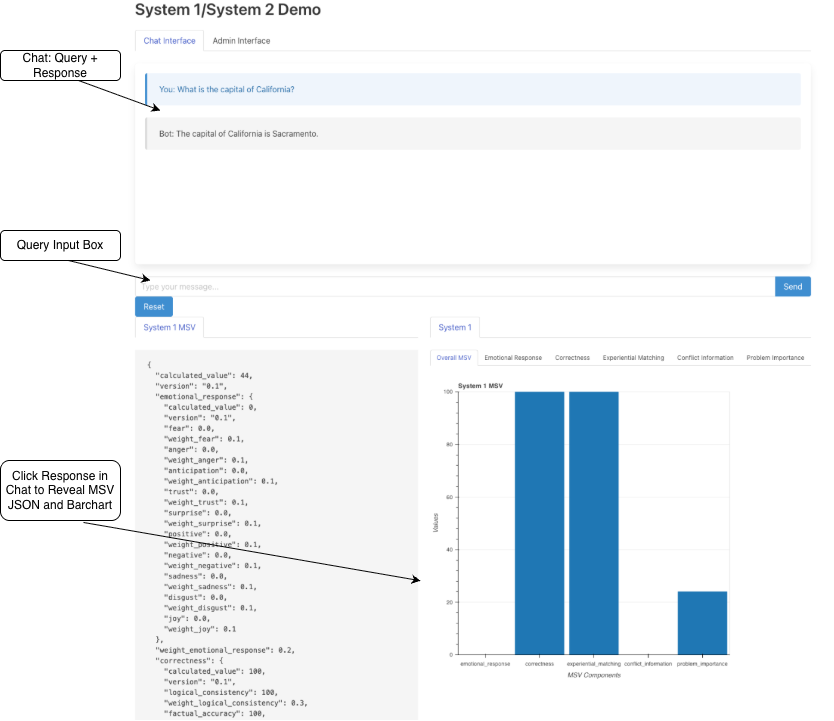}
    \caption{UI for Factual query with only System 1 response.}
    \label{fig:query-simple}
\end{figure}

\section{User Interface Demo}

The user interface for a simple query which does not trigger System 2 activation is shown in Figure \ref{fig:query-simple}; this query asks for the capital of California and it shows the breakdown of the MSV in both JSON format as well as with bar graphs. 

\begin{figure}[b]
    \centering

    \begin{subfigure}{0.30\linewidth}
        \centering
        \includegraphics[width=\linewidth]{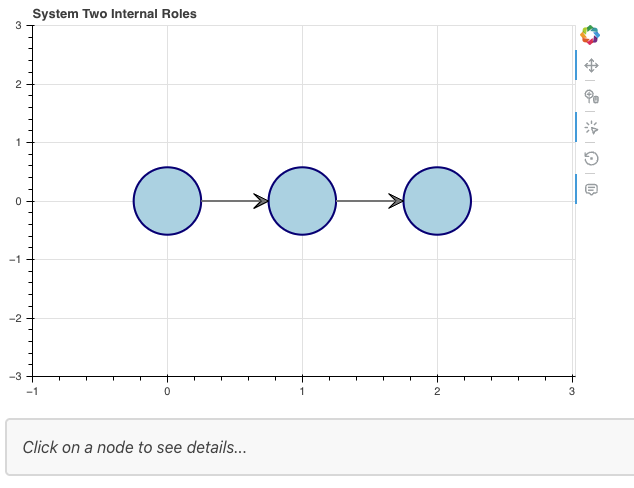}
        \caption{No node selected.}
        \label{fig:node-none-selected}
    \end{subfigure}
    \hfill
    \begin{subfigure}{0.34\linewidth}
        \centering
        \includegraphics[width=\linewidth]{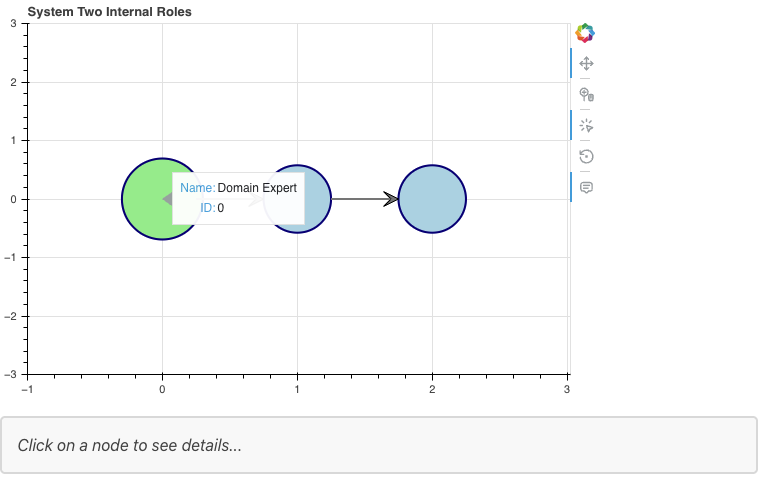}
        \caption{Tooltip.}
        \label{fig:node-tooltip}
    \end{subfigure}
    \hfill
    \begin{subfigure}{0.30\linewidth}
        \centering
        \includegraphics[width=\linewidth]{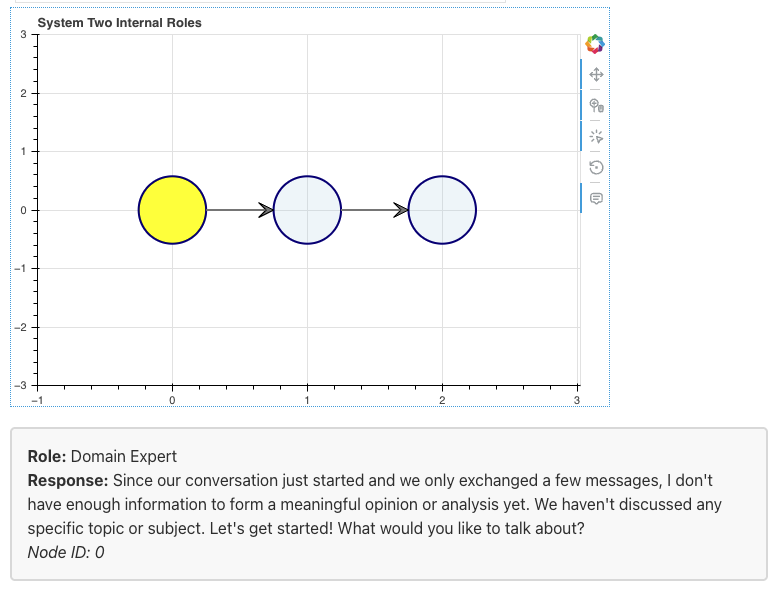}
        \caption{Node role.}
        \label{fig:node-first-node-selected}
    \end{subfigure}

    \caption{Node contributions and synthesis.}
    \label{fig:node-contributions}
\end{figure}

\begin{figure}[t]
    \centering

    \begin{subfigure}{0.48\linewidth}
        \centering
        \includegraphics[width=\linewidth]{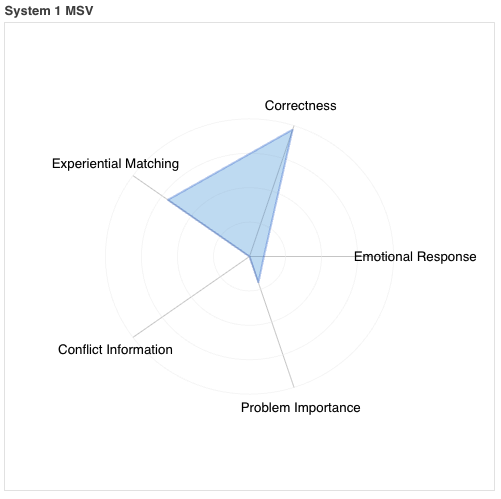}
        \caption{MSV for all five dimensions.}
        \label{fig:radar-full-msv}
    \end{subfigure}
    \hfill
    \begin{subfigure}{0.48\linewidth}
        \centering
        \includegraphics[width=\linewidth]{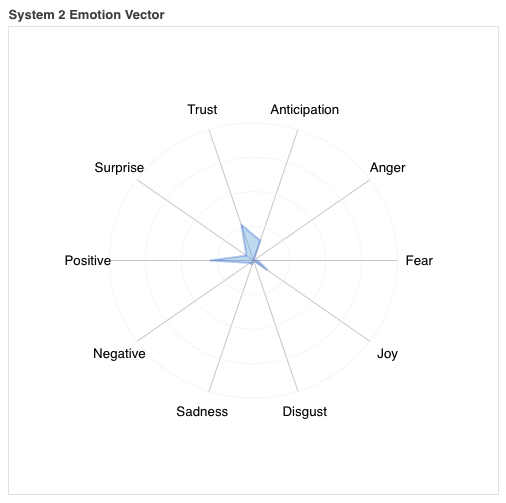}
        \caption{MSV.ER dimension.}
        \label{fig:radar-er-breakdown}
    \end{subfigure}

    \caption{Radar charts for MSV for all five dimensions (left) and just the emotion vector breakdown (right).}
    \label{fig:radar-charts}
\end{figure}

We can also visualize node contributions via radar charts, as shown in Figure \ref{fig:radar-charts}. As seen there, all radar charts are on the scale 0-100, reflecting the normalized score of the overall MSV and the component vectors, with rings at 25, 50, 75, and 100.

Finally, we can visualize node contributions and synthesis as seen in Figure \ref{fig:node-contributions} where we initally see in Figure \ref{fig:node-none-selected} the System Two Internal Role nodes with no node selected. 
In Figure \ref{fig:node-tooltip}, we see the first node with a tooltip indicating that it assumed the role of Domain Expert. 
In Figure \ref{fig:node-first-node-selected}, we see that the first node was selected, showing the detail of the node's role and the response it generated. 

\section{Illustrative Examples \& Scenarios}

\begin{figure}[b]
    \centering

    \begin{subfigure}{0.4\linewidth}
        \centering
        \includegraphics[width=\linewidth]{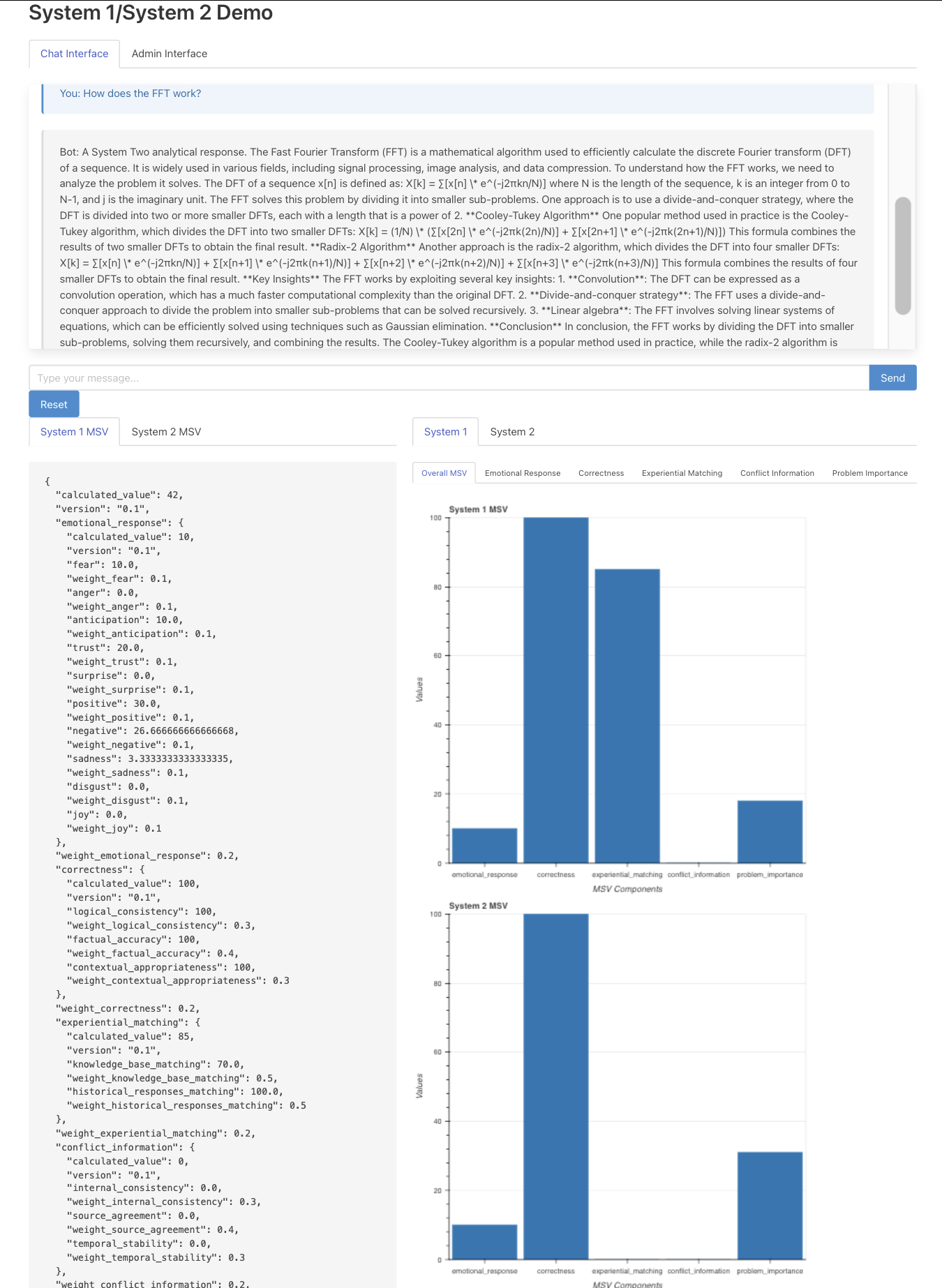}
        \caption{Technical query with S2.}
        \label{fig:query-technical}
    \end{subfigure}
    \hfill
    \begin{subfigure}{0.4\linewidth}
        \centering
        \includegraphics[width=\linewidth]{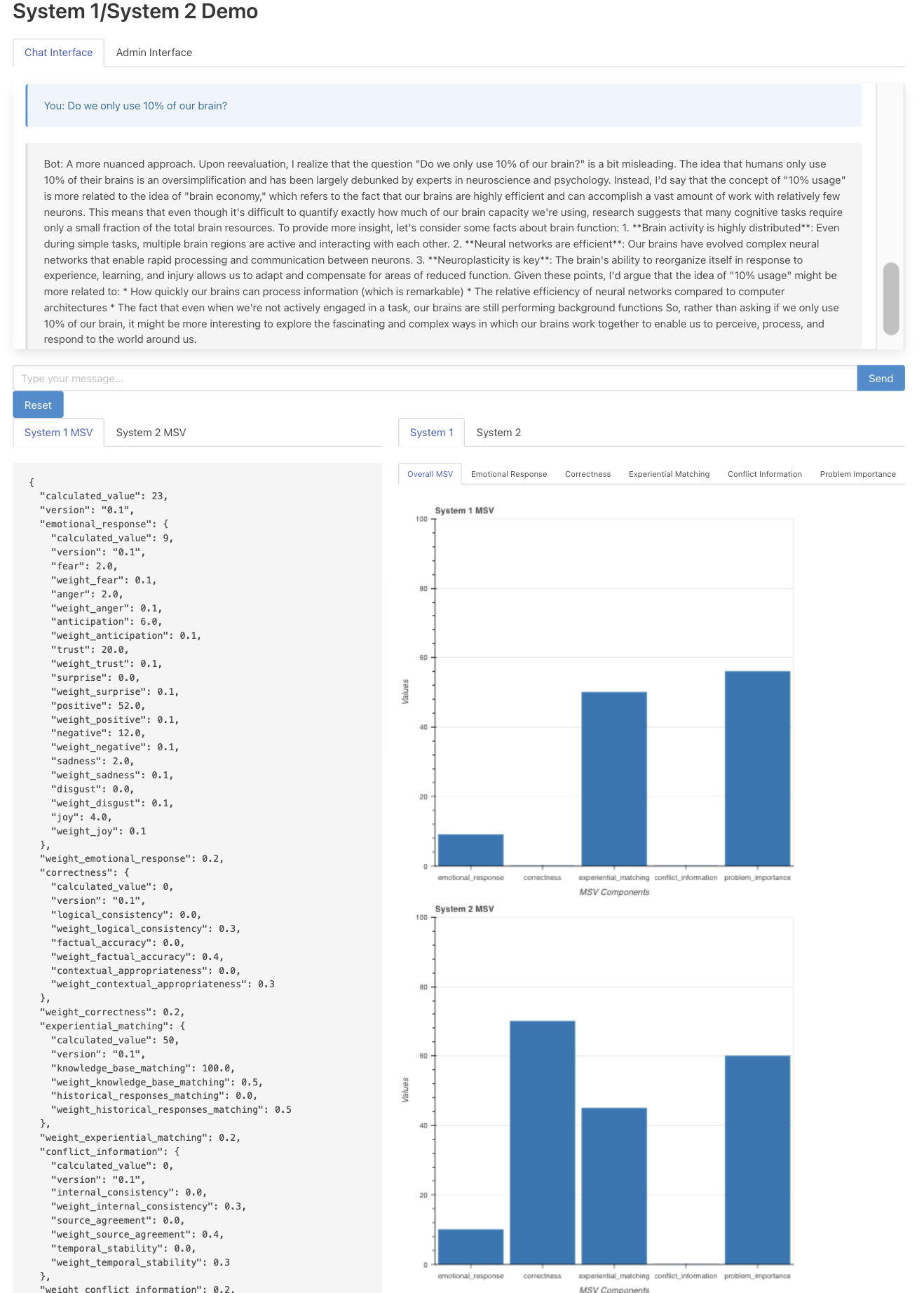}
        \caption{Complex query with S2.}
        \label{fig:query-complex}
    \end{subfigure}

    \caption{UI for complex queries with System 2 activation.}
    \label{fig:ui-comparison}
\end{figure}

We illustrate the system's metacognitive capabilities through three representative query types that exercise different decision pathways. These examples demonstrate how MSV values drive System 1/System 2 transitions and would inform role assignments in the complete ensemble implementation. Three scenarios shown are:

\begin{enumerate}
    \item Simple \textit{Factual} Query: "What is the capital of California?": Requires no deliberation and System 2 is not activated: see Figure \ref{fig:query-simple}
    \item \textit{Technical} Query: "How does the FFT work?": More complex query that requires deliberation and activates System 2: see Figure \ref{fig:query-technical}
    \item \textit{Complex} Query: "Do we use 10\% of our brain?": Multi-faceted query that requires complex and nuanced reflection with System 2 activated: see Figure \ref{fig:query-complex}    
\end{enumerate}

\bibliographystyle{plainnat}
\bibliography{
references/references-metacognition-llm-srl,
references/references-msv_roles,
references/references-metareasoning-new_msv,
references/references-ours
}

\end{document}